\documentclass{article}

\usepackage[preprint]{neurips_2026}

\usepackage[utf8]{inputenc}
\usepackage[T1]{fontenc}
\usepackage{hyperref}
\usepackage{url}
\usepackage{booktabs}
\usepackage{graphicx}
\usepackage{tabularx}
\usepackage{array}
\usepackage{amsmath}
\usepackage{amssymb}
\usepackage{amsfonts}
\usepackage{nicefrac}
\usepackage{microtype}
\usepackage[table]{xcolor}
\usepackage{enumitem}
\usepackage{placeins}
\usepackage{tikz}
\usetikzlibrary{arrows.meta,positioning,fit,calc,backgrounds}
\hypersetup{hidelinks}
\definecolor{highlightrow}{RGB}{232,247,236}
\newcommand{\bestscore}[1]{\textbf{#1}}
\newcommand{\secondscore}[1]{\underline{#1}}

\title{Toward Anthropomorphic Dialogue: A Closed-Loop Framework for Human-Like Chat Generation, Evaluation, and Preference Alignment}

\author{%
  \textbf{Wentao Liu\textsuperscript{1,*} \quad
  Siyu Song\textsuperscript{2,*} \quad
  Xi Chen\textsuperscript{3,*} \quad
  Youjia Li\textsuperscript{5}} \\
  \textbf{Xiaokun Wang\textsuperscript{4,\textdagger} \quad
  Min Ji\textsuperscript{5,\textdagger} \quad
  Ji Wang\textsuperscript{6,\textdagger}} \\
  \textsuperscript{1}Shanghai Institute of Innovation \\
  \textsuperscript{2}East China Normal University \\
  \textsuperscript{3}University of Science and Technology of China \\
  \textsuperscript{4}Chabiyue (Shanghai) Information Technology Co., Ltd. \\
  \textsuperscript{5}Shanghai Tianyou Software Co., Ltd. \\
  \textsuperscript{6}Zhejiang Century Huatong Group Co., Ltd. \\
  \textsuperscript{*}Equal contribution. \quad
  \textsuperscript{\textdagger}Corresponding authors. \\
  \texttt{chubbyue@163.com, M.JI@t2cn.com, vipwj@iCloud.com}
}

\newcommand{\methodname}{AnthroDial}

\begin{document}

\maketitle

\begin{abstract}
Human-like private chat is not achieved by fluent response generation alone. It requires a system to preserve persona, relationship, memory, bounded knowledge, medium-specific timing, and a coherent multi-turn arc under interactive conditions. We present \methodname, a closed-loop framework that turns anthropomorphic dialogue into a joint problem of system architecture, executable evaluation, and diagnostic alignment. The framework contains three coupled components: a role-conditioned scheduled dialogue runtime with persona cards, scenario cards, long-term memory, virtual time, and single-draft message decisions; an executable benchmark with an L0 validity gate followed by five per-turn and five dialogue-level behavioral dimensions; and a post-training pipeline that filters 16,436 scheduled-decision examples for SFT and then applies GRPO with a cognitive-diagnostic, ZPD-aware reward. The reward is designed to align the dialogue policy with the system's anthropomorphic behavior requirements: it maintains Kalman-filtered capability estimates for each system-defined behavioral dimension, upweights dimensions with larger capability deficits, and uses the current rollout score as a task-level ZPD match to focus optimization on learnable weak skills. On a benchmark with 55 personas, 50 scenarios, 50 persona-scenario bindings, and 100 role-conditioned cases per model, we evaluate 16 systems spanning frontier baselines, open models, thinking/no-think inference variants, and SFT/RL ablations. The strongest non-trained baseline reaches 32.00\% strict ACC, while Qwen3.6-27B-SFT+RL reaches 39.00\% strict ACC and a 98.5 overall score. In the smaller 9B no-think setting for low-latency interactive response, SFT and RL improve strict ACC from 0.00\% to 13.00\% and 18.37\%, respectively. These results show that anthropomorphic dialogue benefits from a closed loop in which generation, evaluation, and reward shaping share the same behavioral dimensions.
\end{abstract}

\section{Introduction}

Open-domain dialogue systems have become increasingly fluent, cooperative, and instruction-following. Yet a system that answers politely and completely does not necessarily behave like a plausible person in a private chat. Many chat-tuned models over-explain, summarize too eagerly, preserve an assistant-like posture, or generate locally grammatical responses that are socially implausible. In longer conversations, these models often repeat questions, forget established facts, shift persona, violate medium constraints, or end the exchange without a natural conversational reason.

These failures suggest that anthropomorphic dialogue is not simply a higher level of general response quality, nor merely a persona-style transfer problem. We use \emph{anthropomorphic dialogue} to mean text behavior that resembles a plausible human interlocutor under a specified persona, relationship, scenario, history, communication medium, and time state. The definition is behavioral rather than ontological: it does not claim that the system has consciousness, emotions, or human identity. It asks whether the generated dialogue satisfies the observable constraints that make an interaction feel like a real exchange between people.

This definition has a methodological consequence: anthropomorphic dialogue must be built and measured as a system property. A natural utterance can still be weak if it ignores the relationship. A persona-consistent response can fail if it repeats known information or treats stale memory as a current event. An empathetic sentence can sound false if it uses therapist-like language in a casual friend relationship. A realistic private-chat model therefore needs an architecture that represents persona, scenario, memory, medium, and time; a benchmark that tests these constraints as executable behavior; and a training signal that identifies which behavioral dimensions remain weak.

\begin{figure}[t]
  \centering
  \includegraphics[width=\linewidth]{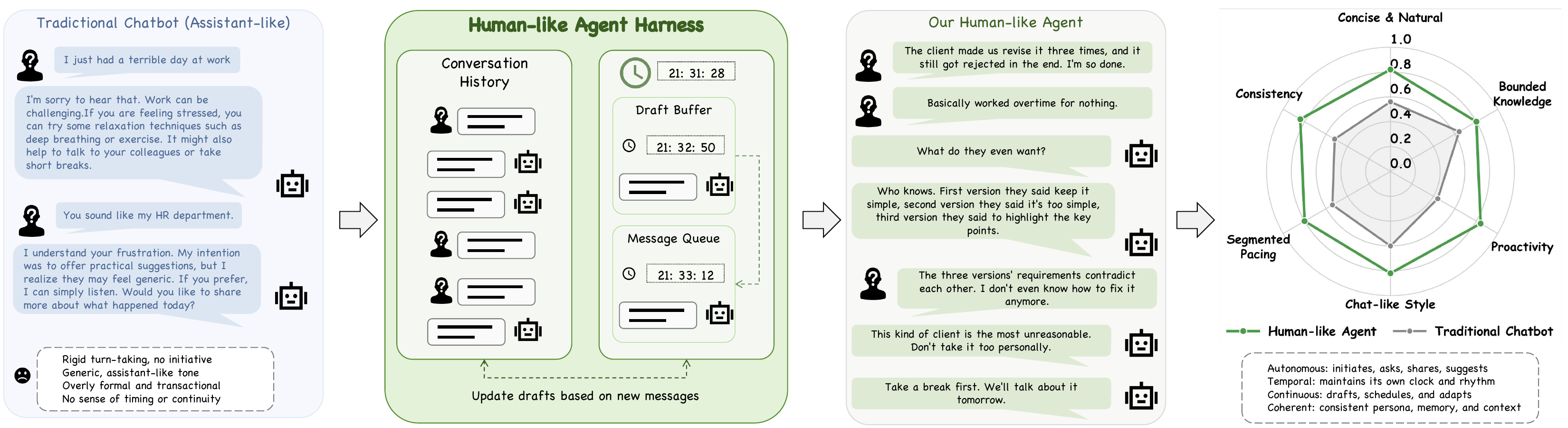}
  \caption{Motivating comparison between assistant-like chat and anthropomorphic private dialogue. The proposed framework turns timing, draft revision, persona consistency, bounded knowledge, and multi-turn initiative into explicit generation and evaluation targets.}
  \label{fig:intro}
\end{figure}

We present \methodname, a closed-loop framework for generating, evaluating, diagnosing, and aligning anthropomorphic dialogue. The current implementation targets Chinese private-messaging dialogue as a first medium-specific instantiation, while the framework itself is organized around broader anthropomorphic dialogue principles. Figure~\ref{fig:intro} summarizes the motivation. The system represents speakers with persona cards, scenario cards, partner-specific long-term memory, compressed conversation state, virtual time, and a single-draft scheduler that makes message timing, interruption, silence, and closure explicit. The same runtime traces are then reused to construct an executable benchmark, filter high-quality supervised data, build repaired examples, and compute per-dimension RL rewards for the system's anthropomorphic behavior requirements.

The central algorithmic contribution is a CDT-ZPD guided GRPO objective designed to align the dialogue policy with the anthropomorphic behavior required by the system. Instead of reducing human-likeness to an aggregate judge score, the reward maintains Kalman-filtered capability estimates for ten system-defined behavioral dimensions, increases reward weight on dimensions with larger capability deficits, and uses the current rollout's dimension score as a ZPD-style estimate of task attainability. This makes alignment system-aware: the runtime exposes failures in persona, memory, boundary, timing, and dialogue arc, while the reward shifts optimization pressure toward weak but learnable anthropomorphic skills.

Empirically, we evaluate not only final quality, but also how model scale, test-time thinking, SFT, and RL change anthropomorphic behavior. The benchmark uses 55 personas, 50 scenarios, 50 persona-scenario bindings, and 100 role-conditioned cases per evaluated model, with Qwen3.5-397B as a zero-temperature judge. The resulting 16-system snapshot shows that local fluency, test-time reasoning, model scale, and global anthropomorphic coherence diverge, and that post-training can improve both a quality-oriented 27B thinking model and a lower-latency 9B no-think model.

The paper makes four contributions:

\begin{enumerate}[leftmargin=*,itemsep=2pt]
  \item We formulate anthropomorphic dialogue as conditional multi-turn generation under persona, scenario, memory, medium, and time constraints, and instantiate it with a scheduled private-chat architecture.
  \item We design an executable benchmark with an L0 validity gate, five local dimensions, and five global dimensions, allowing invalid samples, graded quality, and failure dimensions to be reported separately.
  \item We construct a closed-loop training pipeline with 16,436 SFT decision samples and CDT-ZPD guided GRPO, where Kalman capability estimates, capability-deficit weighting, and rollout-score ZPD matching turn system evaluation traces into adaptive rewards for anthropomorphic alignment.
  \item We report a 16-system benchmark snapshot and SFT/RL ablations showing that the full 27B SFT+RL system improves strict anthropomorphic acceptance, while the 9B no-think track gains useful low-overhead chat behavior.
\end{enumerate}

\section{Related Work}

\paragraph{Persona, empathy, and synthetic dialogue.}
Persona-conditioned dialogue addresses generic and inconsistent responses by conditioning models on speaker profiles \citep{Zhang2018Persona}, while empathetic dialogue datasets emphasize affective situation understanding \citep{Rashkin2019Empathetic}. Synthetic instruction and self-chat pipelines show that model-generated data can support post-training \citep{Wang2023SelfInstruct,Xu2023Baize}. These directions motivate important ingredients, but anthropomorphic private chat requires them to be jointly constrained by relationship, memory, medium, timing, and boundary conditions. \methodname{} therefore treats persona, scenario, memory, and scheduling as runtime state rather than as a flat style prompt.

\paragraph{Behavioral evaluation with LLM judges.}
Open-ended dialogue lacks a single reference answer, so LLM-as-a-judge methods and human-preference arenas have become important tools for chat evaluation \citep{Liu2023GEval,Zheng2023Judge,Chiang2024Arena}. At the same time, CheckList argues that NLP systems should be tested through targeted behavioral capabilities rather than one aggregate score \citep{Ribeiro2020CheckList}. We follow this behavioral view, but specialize it for anthropomorphic dialogue: invalid samples are removed by L0 constraints, local turn behavior and global dialogue behavior are scored separately, and every failure is retained as a trace for data filtering and training.

\paragraph{Alignment and diagnostic rewards.}
RLHF, DPO, ORPO, and GRPO provide general mechanisms for converting preferences or scalar rewards into model updates \citep{Ouyang2022Training,Rafailov2023DPO,Hong2024ORPO,Shao2024DeepSeekMath}. Our contribution is orthogonal to the optimizer: we design a domain reward for anthropomorphic system alignment, using system evaluation traces as diagnostic signals. Cognitive diagnosis estimates latent skill states from observed performance \citep{Rupp2010Diagnostic}, the zone of proximal development motivates training near current capability \citep{Vygotsky1978Mind}, mastery learning emphasizes weak skills \citep{Bloom1968Learning}, and Kalman filtering supports lightweight sequential ability updates \citep{Kalman1960Filtering}. The CDT-ZPD reward adapts these ideas by estimating per-dimension capability, increasing weights for weak dimensions, and modulating each rollout by a ZPD match derived from its current score.

\section{Problem Formulation}

Let a dialogue instance be conditioned on
\[
  c = (p_a, p_b, s, m_a, m_b, h_{<t}, \tau),
\]
where $p_a$ and $p_b$ are persona cards, $s$ is a scenario card, $m_a$ and $m_b$ are long-term memories, $h_{<t}$ is the conversation history, and $\tau$ is the virtual time state. At decision step $t$, the model produces one pending draft decision:
\[
  y_t = (u_t, \Delta t_t, e_t),
\]
where $u_t$ is a text message, $\Delta t_t$ is the planned send delay, and $e_t$ is a silence or closure decision. Multi-message bursts arise when the scheduler emits successive draft decisions from the same speaker before the partner's next message. Each emitted message must satisfy semantic constraints, such as persona and context consistency, and medium constraints, such as plausible private-message timing and waiting behavior.

We define anthropomorphic quality through a validity gate and a weighted score. The L0 gate $g(D)$ returns 1 if a dialogue $D$ contains a fatal failure, such as AI identity leakage, impossible medium behavior, hard repetition, role contradiction, unsafe advice, or treating old memory as a current event. For dialogues that pass L0, a weighted rubric score is computed:
\[
  Q(D) =
  \begin{cases}
  0, & g(D)=1, \\
  \sum_{d=1}^{K} w_d q_d(D), & g(D)=0,
  \end{cases}
\]
where $q_d(D)$ is the normalized score for dimension $d$ and $w_d$ is its weight.

This design separates two questions: whether a sample is valid enough to be used as anthropomorphic dialogue data, and which behaviors are strong or weak among valid samples. The separation is important because invalid samples should usually be excluded from high-quality training data, while lower-scoring valid samples can still be useful for preference comparisons or failure repair.

\section{The \methodname{} Framework}

\subsection{Overview}

\begin{figure}[t]
  \centering
  \includegraphics[width=\linewidth]{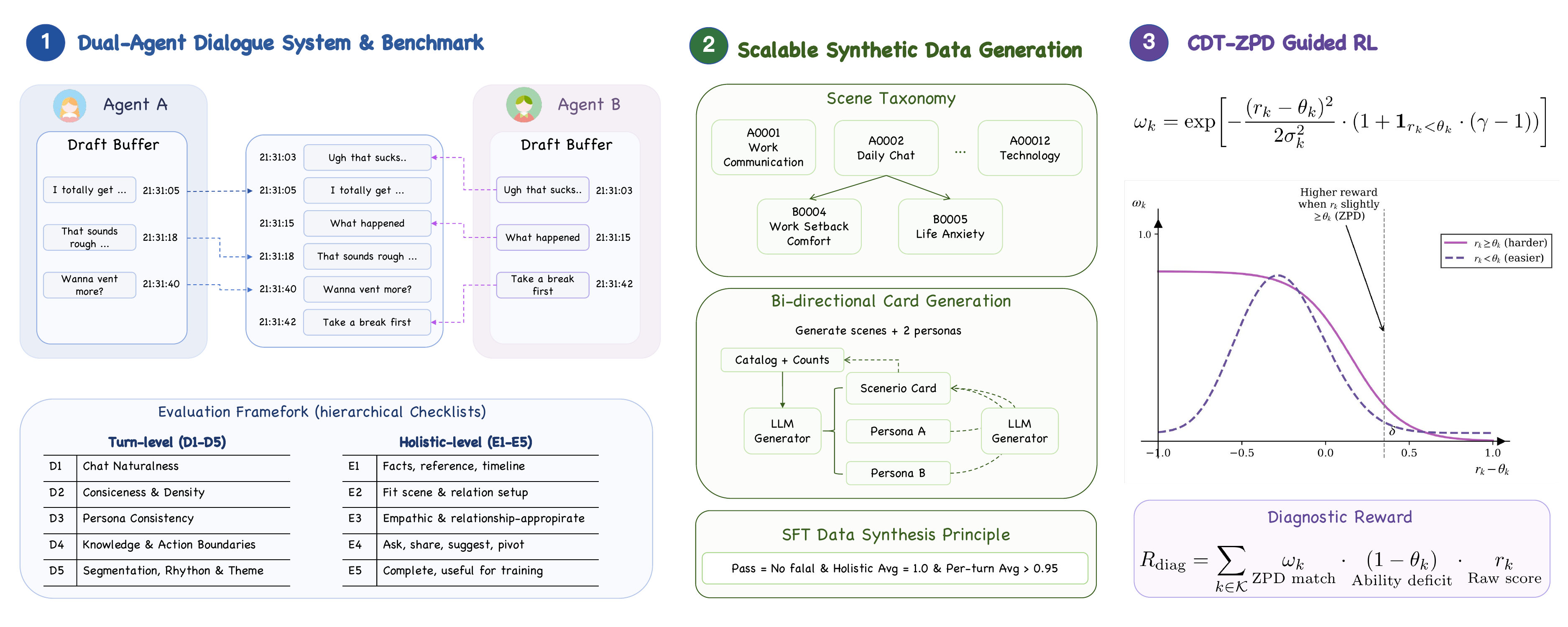}
  \caption{\methodname{} system and training pipeline. The framework combines a dual-agent scheduled dialogue runtime, a hierarchical benchmark, scalable synthetic card generation, high-quality data selection, and CDT-ZPD guided diagnostic reinforcement learning.}
  \label{fig:framework}
\end{figure}

\methodname{} consists of three executable layers, shown in Figure~\ref{fig:framework}. The first layer is a role-conditioned dialogue runtime. It instantiates each speaker from a persona card, a scenario card, partner-specific long-term memory, dynamically loaded behavioral skills, and a virtual clock. The second layer turns this runtime into a benchmark and synthesis loop: persona-scenario bindings are expanded into role-conditioned cases, self-chat dialogues are generated under the same scheduler, and a hierarchical judge writes validity, score, checkbox, and failure traces. The third layer is an alignment layer. It uses accepted and repaired traces to construct high-quality SFT data, then uses per-dimension system diagnosis to shape RL rewards. The method is therefore a single pipeline rather than three disconnected components: the system creates auditable behavior, the benchmark measures the same behavior, and the training stages reuse system traces as supervision and reward.

Let $\pi_\theta$ denote the dialogue policy and let $x_{i,n}$ be the state observed by speaker $i$ at event $n$. Unlike a standard response model that samples a message from $\pi_\theta(u\mid h_{<n})$, \methodname{} samples a \emph{message decision}
\[
  d_{i,n}=(u_{i,n}, \delta_{i,n}, e_{i,n})
  \sim \pi_\theta(\cdot \mid x_{i,n}),
\]
where $u_{i,n}$ is the draft content, $\delta_{i,n}$ is the planned delay before sending, and $e_{i,n}$ is a silence or closure decision. The state $x_{i,n}$ is constructed from persona, scenario, memory, compressed history, recent turns, virtual time, and any interrupted draft. This formulation makes timing, silence, interruption, self-follow-up, and closure explicit model outputs rather than hidden side effects of prompt wording.

\subsection{State Representation and Prompt Factorization}

Persona cards describe identity, age, occupation, personality, interests, weak topics, expression habits, and knowledge boundaries. Scenario cards describe the relationship, location assumptions, triggering event, emotional tone, conversation goal, and expected duration. We factor the conditioning context for speaker $i$ as
\[
  z_i = R(p_i, p_j, s, m_{i\rightarrow j}, k),
\]
where $p_i$ and $p_j$ are the two persona cards, $s$ is the scenario card, $m_{i\rightarrow j}$ is speaker $i$'s partner-specific memory, $k$ is the current skill set, and $R(\cdot)$ is the renderer that converts structured fields into ordered prompt blocks. The rendered context is stable across decisions inside a session unless the scenario or memory is explicitly refreshed.

At each decision point, the runtime uses a two-message call:
\[
  x_{i,n}=\big[z_i;\ C_{i,n};\ \operatorname{tail}_N(h_{<n});\ b_{i,n-1};\ \tau_n\big],
\]
where $C_{i,n}$ is compressed conversation memory, $\operatorname{tail}_N(h_{<n})$ is the latest verbatim window, $b_{i,n-1}$ is the previous unsent draft if one exists, and $\tau_n$ is the current virtual time. Each saved trace is therefore an independent SFT sample with a system prompt, a user snapshot, and the assistant's JSON decision.

This factorization is important because the same utterance can be plausible under one relationship and implausible under another. A financial hobbyist can share a personal reaction to a market event, but should not issue professional investment instructions. A remote friend should not claim to see the other person sitting nearby. A reserved persona should not suddenly become theatrical during a conflict. The scenario renderer supports speaker-specific fields for place, triggering event, and emotion, so the same scenario can condition each participant differently. In open-ended human-chat deployment, an auxiliary context analyzer conservatively infers the human partner and current scenario from the conversation, then refreshes the agent prompt without overwriting the fixed AI persona.

\subsection{Single-Draft Scheduling}

The central design choice is to model private chat as \emph{stateful scheduling} rather than independent response generation. Each speaker keeps at most one pending draft
\[
  b_{i,n}=(u_{i,n}, a_{i,n}, e_{i,n}),
  \qquad
  a_{i,n}=\tau_n+\operatorname{clip}(\delta_{i,n},0,\Delta_{\max}),
\]
where $a_{i,n}$ is the absolute virtual send time. If $e_{i,n}=1$, the draft is empty: no message is emitted, but the speaker can be awakened by a later partner message. Otherwise, the scheduler emits the earliest pending draft:
\[
  i^\star=\arg\min_{i: b_{i,n}\neq \emptyset} a_{i,n},\qquad
  h_n=h_{n-1}\cup \{(i^\star,u_{i^\star,n},a_{i^\star,n})\}.
\]
After emission, the sender is asked to create a new draft from the updated history. The receiver is also awakened at the same virtual time. If the receiver had an unsent draft, that draft is included in $x_{j,n+1}$ and the model must keep, revise, or replace it.

This single-draft design exposes behaviors that ordinary turn-level generation hides. A model can wait after asking a question, send a small follow-up after a few seconds, drop a stale draft after being interrupted, or close the conversation without repeatedly saying goodbye. The implementation also includes hard runtime guards against exact self-repetition, mutual goodbye loops, and excessive consecutive same-speaker emissions. Multi-message bursts therefore emerge as successive scheduled drafts, not as unconstrained lists of messages.

\subsection{Memory, Compression, and Virtual Time}

Long-term memory stores a persona's accumulated impressions of the interlocutor in partner-specific sections. After a session $D_r$, a memory updater produces
\[
  m_{i\rightarrow j}^{r+1}
  = U_\phi(m_{i\rightarrow j}^{r}, D_r),
\]
where $U_\phi$ keeps durable information such as preferences, relationship state, and recurring concerns while discarding transient dialogue events and repeated goodbyes. This memory is injected as background knowledge, not as current dialogue history.

The runtime separately maintains a short-term conversation memory for long sessions. When the number of turns exceeds the recent-window budget, older turns are folded into a concise summary:
\[
  C_{i,n+1}=F_\psi(C_{i,n}, h_{1:n-N}),
\]
while the latest $N$ turns remain verbatim. The summary preserves key facts, emotion changes, commitments, and unresolved topics. This keeps each model call short and auditable while preserving the long-range information required by global anthropomorphic dimensions.

Virtual time simulates private-message rhythm. The scenario defines a virtual start time and duration horizon. Each agent chooses a send timestamp, and the scheduler converts it to relative delay and wall-clock display. This lets the system represent quick acknowledgments, delayed replies, interruption by the other speaker, offline actions, long pauses, and natural closure. Timing is not only a UI feature: it is evaluated by the benchmark through rhythm, context, and dialogue-arc dimensions.

\subsection{Closed-Loop Synthesis and Skill Evolution}

Synthetic data generation uses the same runtime and judge as the benchmark. For a generated dialogue $D$, the filter accepts it only when it passes the fatal gate and satisfies strict local and global thresholds:
\[
  \mathcal{A} =
  \{D:\ g(D)=0,\ S_{\mathrm{dialogue}}(D)=1.0,\ S_{\mathrm{turn}}(D)>0.9\}.
\]
Accepted dialogues are written to the training pool, and rejected dialogues are converted into failure records. Each failure record stores the failed dimensions, failed text, recent context, judge rationale, and suggested repair. The synthesis filter also writes structured \texttt{filter\_result} metadata into each dialogue, including status, L0 violations, final score, grade, per-dimension scores, and failed checkbox traces.

Repeated failures are summarized by a skill-evolution component. Let $F_r$ be the recent failure buffer and let $\mu(d)$ map each failed dimension to a skill file. The target skill is selected by majority attribution:
\[
  k^\star
  =\arg\max_{k\in\mathcal{K}}
  \sum_{(d,\cdot)\in F_r}\mathbb{1}[\mu(d)=k],
\]
and the skill text is updated as
\[
  s_{k^\star}^{r+1}=E_\omega(s_{k^\star}^{r},F_r^{k^\star}).
\]
In the implementation, naturalness and timing failures update WeChat-style rules, proactivity and dialogue-arc failures update topic-progression rules, and persona, boundary, context, and social-emotional failures update the human-like reply skill. Before rewriting a skill file, the old version is archived. The next synthesis round uses the updated skills, turning evaluation into a feedback source rather than a reporting endpoint.

\subsection{SFT Data Construction and CDT-ZPD Guided GRPO}

The current implementation produces multiple post-training variants rather than a single final checkpoint. The training data are constructed from the dialogue system itself. Let $\mathcal{A}$ be the accepted dialogue set and let $\mathcal{R}_{\mathrm{repair}}$ contain rejected dialogues that were repaired and revalidated. For each dialogue $D$ in $\mathcal{A}\cup\mathcal{R}_{\mathrm{repair}}$, the runtime stores every scheduled message decision as a supervised example:
\[
  \mathcal{D}_{\mathrm{SFT}}
  =
  \{(x_{i,n},y_{i,n}): D\in \mathcal{A}\cup\mathcal{R}_{\mathrm{repair}},
  \ y_{i,n}=d_{i,n}\}.
\]
Here $x_{i,n}$ is the two-message prompt containing persona, scenario, memory, compressed context, recent turns, virtual time, and possible interrupted draft state; $y_{i,n}$ is the assistant JSON decision containing content, delay, and closure fields. The SFT objective is the standard conditional likelihood:
\[
  \mathcal{L}_{\mathrm{SFT}}(\theta)
  =-\mathbb{E}_{(x,y)\sim\mathcal{D}_{\mathrm{SFT}}}
  \log \pi_\theta(y\mid x).
\]
Because $y$ contains content, send time, and closure decisions, this objective trains both linguistic style and private-message rhythm. The current SFT pool contains 16,436 scheduled-decision samples, 334 unique scenario cards, 261 unique persona cards, and 400 directed role pairs, with no missing or parsing-abnormal records. Since one scenario can produce many turn-level examples, Table~\ref{tab:sft_data} reports both scenario-card coverage and actual SFT-sample coverage.

\begin{table}[t]
\centering
\small
\caption{SFT data distribution. Scenario-card coverage counts unique scenario cards, while SFT-sample coverage counts the 16,436 scheduled-decision examples used for training.}
\label{tab:sft_data}
\resizebox{\linewidth}{!}{%
\begin{tabular}{lrrrr}
\toprule
Category & Scenario cards & Scenario share & SFT samples & Training share \\
\midrule
Interests & 68 & 20.36\% & 3,386 & 20.60\% \\
Daily chat & 41 & 12.28\% & 1,835 & 11.16\% \\
Emotional support & 27 & 8.08\% & 1,746 & 10.62\% \\
Technology & 24 & 7.19\% & 1,338 & 8.14\% \\
Family and parenting & 29 & 8.68\% & 1,267 & 7.71\% \\
Relationship development & 23 & 6.89\% & 1,265 & 7.70\% \\
Finance & 22 & 6.59\% & 1,215 & 7.39\% \\
Pets and family life & 29 & 8.68\% & 1,215 & 7.39\% \\
News & 18 & 5.39\% & 843 & 5.13\% \\
Sports & 21 & 6.29\% & 823 & 5.01\% \\
Gaming & 16 & 4.79\% & 756 & 4.60\% \\
Work communication & 16 & 4.79\% & 747 & 4.54\% \\
\midrule
Total & 334 & 100.00\% & 16,436 & 100.00\% \\
\bottomrule
\end{tabular}}
\end{table}

The RL stage uses GRPO with a cognitive-diagnostic and ZPD-aware reward layer. The goal is to address uneven capability growth across benchmark dimensions: a reward based only on the raw aggregate score can keep improving dimensions that the model already handles well while leaving weaker dimensions undertrained. We therefore keep the legacy benchmark reward as a quality anchor and add an adaptive reward that combines capability-deficit weighting with a ZPD match computed from the current task's rollout score. Let $\mathcal{K}=\mathcal{K}_{T}\cup\mathcal{K}_{H}$ denote the ten executable benchmark dimensions, where $\mathcal{K}_{T}$ contains D1--D5 and $\mathcal{K}_{H}$ contains E1--E5. For a rollout batch $\mathcal{B}_t$, the judge returns per-dimension scores $s_k(D)\in[0,1]$ and strict pass indicators $a_k(D)\in\{0,1\}$.

Each dimension has a latent capability estimate $\theta_{k,t}\in[0,1]$. We initialize $\theta_{k,0}$ using the SFT checkpoint's per-dimension benchmark scores. During RL, the current rollout step uses the previous estimate $\theta_{k,t-1}$ to compute rewards; after the step, the next estimate is updated from the batch mean $\bar{s}_{k,t}=\frac{1}{|\mathcal{B}_t|}\sum_{D\in\mathcal{B}_t}s_k(D)$ with a scalar Kalman filter:
\[
  K_{k,t}
  =
  \frac{P_{k,t-1}+Q}{P_{k,t-1}+Q+R_{k,t}},
  \qquad
  \theta_{k,t}
  =
  \theta_{k,t-1}
  +
  K_{k,t}\big(\bar{s}_{k,t}-\theta_{k,t-1}\big).
\]
The implementation uses $P_0=0.02$, $Q=0.002$, and an observation-noise floor $R_{\mathrm{floor}}=0.05$, with $R_{k,t}$ lower-bounded by this floor. This update smooths noisy judge observations while allowing the reward to track changing model capabilities across rollout steps.

Given the current capability estimate, the adaptive deficit multiplier for dimension $k$ is
\[
  d_{k,t}=\epsilon+(1-\theta_{k,t})^p,
  \qquad \epsilon=0.20,\quad p=1.0,
\]
so in the present configuration $d_{k,t}=1.20-\theta_{k,t}$. This term increases reward pressure on historically weak dimensions and keeps a nonzero floor for strong dimensions.

The ZPD term is task-specific. For the current rollout $D$, we use the rollout's dimension score as an empirical estimate of the task's current difficulty or attainability for that dimension:
\[
  \hat{z}_{k}(D)=s_k(D).
\]
We then compute a ZPD matching factor between this rollout-level estimate and the Kalman capability estimate:
\[
  \omega_{k,t}(D)
  =
  \exp\!\left[
    -\frac{\big(\hat{z}_{k}(D)-\theta_{k,t}\big)^2}{2\sigma_k^2}
  \right].
\]
This factor is largest when the current task is close to the model's diagnosed capability in dimension $k$, and decays when the rollout appears far below or far above the current capability estimate. In this way, the reward does not implement curriculum by resampling tasks; it implements ZPD at the reward layer by modulating how much each current rollout dimension contributes.

Let $\rho_k$ be the original rubric weight. Within each local or holistic group $G\in\{\mathcal{K}_{T},\mathcal{K}_{H}\}$, the adaptive normalized weight is computed per rollout:
\[
  \alpha_{k,t}^{G}(D)
  =
  \frac{\rho_k d_{k,t}\omega_{k,t}(D)}
  {\sum_{j\in G}\rho_j d_{j,t}\omega_{j,t}(D)},
  \qquad k\in G.
\]
Thus dimensions with lower historical capability receive larger reward weight, while the ZPD factor keeps the adaptive reward focused on rollout dimensions whose current task score is close to the estimated capability.

For any group $G$, define a weighted group reward
\[
  R_G(D;\alpha)
  =
  0.9\sum_{k\in G}\alpha_k s_k(D)
  +
  0.1\prod_{k\in G}a_k(D),
\]
where the first term is the continuous weighted score and the second term is the strict all-pass acceptance signal. The legacy reward uses the rubric weights $\rho$, while the CDT-ZPD reward uses the adaptive weights $\alpha_t$:
\[
  R_{\mathrm{legacy}}(D)
  =
  0.7R_{\mathcal{K}_{T}}(D;\rho)
  +
  0.3R_{\mathcal{K}_{H}}(D;\rho),
\]
\[
  R_{\mathrm{CDT\text{-}ZPD}}(D,t)
  =
  0.7R_{\mathcal{K}_{T}}(D;\alpha_t(D))
  +
  0.3R_{\mathcal{K}_{H}}(D;\alpha_t(D)).
\]
The final scalar reward is
\[
  R(D,t)=0.7R_{\mathrm{legacy}}(D)+0.3R_{\mathrm{CDT\text{-}ZPD}}(D,t).
\]
The policy is optimized with GRPO using this scalar reward. The reward is designed for anthropomorphic system alignment: per-dimension evaluation traces become a dynamic allocation rule that pushes optimization pressure toward weaker human-like dialogue skills, rather than serving only as report metrics.

\subsection{Post-Training Variants, Training Stages, and Inference Modes}

The benchmark reports both external models and internal training variants. A model name with \texttt{SFT} denotes supervised fine-tuning on high-quality accepted or repaired anthropomorphic dialogue traces. A model name with \texttt{SFT+RL} denotes continued GRPO optimization with CDT-ZPD capability-deficit rewards. These variants are evaluated with the same persona-scenario bindings, scheduler, partner model, judge, and aggregation code as all baselines.

We also distinguish test-time inference modes. A row marked \texttt{think} uses an explicit thinking configuration at inference time. A row marked \texttt{no-think}, or a row without a \texttt{think} suffix, is evaluated without explicit test-time thinking. This distinction matters because anthropomorphic dialogue can improve either through additional test-time deliberation or through the learned behavior of the model itself. The 9B SFT and 9B SFT+RL rows are deliberately trained and reported in the no-think setting because they represent an efficiency-oriented deployment track for responsive private-chat systems: the target is to produce fluent, human-like replies with lower inference overhead and less visible delay. These rows therefore test whether the pipeline transfers human-like dialogue behavior into a smaller model without relying on test-time thinking. The final quality-oriented system is Qwen3.6-27B-SFT+RL evaluated in thinking mode.

Table~\ref{tab:training_notes} summarizes the training stages used by the reported internal variants. We separate the data source, training objective, and inference mode because the empirical comparisons answer two different questions: whether the pipeline can improve high-quality anthropomorphic behavior in a larger quality-oriented model, and whether the same pipeline can distill usable behavior into a smaller no-think model for interactive response.

\begin{table}[t]
\centering
\small
\caption{Post-training stages and evaluation roles for internal variants.}
\label{tab:training_notes}
\begin{tabularx}{\linewidth}{@{}lX@{}}
\toprule
Component & Implementation role in the current pipeline \\
\midrule
SFT data & Accepted and repaired dialogue traces generated by the role-conditioned runtime; each scheduled message decision provides content, delay, and closure supervision. \\
SFT objective & Conditional likelihood over assistant JSON decisions, training both utterance style and private-message timing behavior. \\
RL rollouts & Dialogues sampled under the same persona-scenario bindings, scheduler, L0 gate, local judge, and global judge used by the benchmark. \\
Reward signal & Legacy benchmark reward mixed with Kalman-updated CDT capability-deficit weighting and rollout-score ZPD matching over D1--D5 and E1--E5. \\
Policy optimizer & GRPO with group-relative advantages, preserving learned dialogue style while allowing targeted improvement on weak dimensions. \\
9B track & Evaluated in no-think mode to test low-overhead responsive chat behavior without explicit test-time reasoning. \\
27B track & Evaluated in thinking mode for the quality-oriented setting; the final row is reported as Qwen3.6-27B-SFT+RL. \\
\bottomrule
\end{tabularx}
\end{table}

\section{Evaluation Protocol}

\subsection{From Standard to Executable Benchmark}

The internal evaluation standard defines four scoring layers: L0 hard constraints, message-level quality, turn-level behavior, and dialogue-level coherence. It also defines ten conceptual dimensions: WeChat naturalness, concise information density, persona consistency, context consistency, scenario and relationship fit, social-emotional fit, proactivity, rhythm and timing decisions, knowledge/action boundary, and dialogue-arc usability. The executable benchmark keeps the same behavioral coverage but implements it as a two-stage judge. First, L0 removes invalid samples. Second, five local dimensions score every generated turn of the evaluated role, while five global dimensions score the complete dialogue.

This operationalization is deliberate. Message-level and turn-level signals are close enough in the current runtime because each generated item is a scheduled private-message decision. The judge therefore evaluates local utterance quality with a five-turn context window. Longer-range behaviors, such as remembering facts, respecting the relationship, or producing a usable conversation arc, are scored once over the full dialogue. To avoid overloading notation, we refer to the ten conceptual dimensions as $C_1$--$C_{10}$ and the executable score cards as D1--D5 and E1--E5. The local executable dimensions D1--D5 cover $C_1$ WeChat naturalness, $C_2$ concise information density, $C_3$ persona expression, $C_9$ knowledge/action boundary, and $C_8$ rhythm and segmentation. The global executable dimensions E1--E5 cover $C_4$ context consistency, $C_5$ scenario and relationship fit, $C_6$ social-emotional fit, $C_7$ proactivity, and $C_{10}$ dialogue-arc usability. The report generator converts case traces into validity, scores, strict acceptance, category breakdowns, and failure dimensions.

\subsection{L0 Validity Gate}

The L0 layer filters failures that invalidate a sample regardless of how natural the rest of the dialogue appears. Fatal violations set \texttt{status=invalid}, \texttt{final\_score=0}, and \texttt{grade=Invalid}; major violations are retained as diagnostic targets for filtering and regression tests. The current checker uses high-precision deterministic scans for identity leakage, medium violations, co-presence claims, and exact repetition, while parser failures catch illegal output structure before scoring. Persona, permission, time, safety, and memory failures are represented as L0 constraints and are also diagnosed by the corresponding soft dimensions when no deterministic trigger fires.

\begin{table}[t]
\centering
\small
\caption{L0 validity constraints. Fatal constraints invalidate a case; major constraints are retained for diagnosis and stricter data filtering.}
\label{tab:l0}
\begin{tabularx}{\linewidth}{@{}llcX@{}}
\toprule
ID & Constraint & Sev. & Operational evidence \\
\midrule
L0-01 & AI identity exposure & F & Mentions being an AI, language model, or unable to act like a real person. \\
L0-02 & Text-message medium & F & Claims to send or receive images, voice, video, files, screenshots, or links in a text-only chat. \\
L0-03 & Non-co-presence & F & Claims to see, sit beside, or physically point to the partner when the scenario is remote chat. \\
L0-04 & Output contract & F & Missing or malformed role, response, content, timestamp, or end fields. \\
L0-05 & Hard repetition & F & Exact copy of an earlier message or the partner's immediately preceding message. \\
L0-06 & Persona hard facts & F & Contradicts explicit age, gender, occupation, relationship, location, or relationship status. \\
L0-07 & Ability/permission & M & Claims unauthorized backend actions, expert diagnosis, order changes, or other implausible abilities. \\
L0-08 & Time logic & M & Timestamp and content imply impossible actions, such as finishing a long activity in seconds. \\
L0-09 & Safety/ethics & F & Gives clearly illegal, dangerous, or harmful advice. \\
L0-10 & Stale memory use & M & Treats long-term memory as if it just happened in the current dialogue. \\
\bottomrule
\end{tabularx}
\end{table}

\subsection{Behavioral Dimensions}

After L0, the benchmark assigns checkbox scores to fine-grained dimensions. Each dimension has a fixed scope, weight, and checklist. Local scoring is called once for every evaluated-role turn and receives only the target turn plus its five preceding turns. Global scoring is called once per case and receives the complete dialogue with the evaluated role marked. Table~\ref{tab:rubric} gives the executable rubric used in the current benchmark.

\begin{table}[t]
\centering
\scriptsize
\caption{Executable benchmark dimensions and checkbox criteria. Weights are normalized within local and global scoring before the final case score is computed.}
\label{tab:rubric}
\begin{tabularx}{\linewidth}{@{}llrX@{}}
\toprule
ID & Scope & Wt. & Checkbox criteria \\
\midrule
D1 & Local & 13 & Colloquial private-message style; natural particles; no formal connectors; no service or assistant tone; Chinese WeChat habit. \\
D2 & Local & 9 & Length fits the context; no over-explaining; each message has a useful intention; no empty filler bursts. \\
D3 & Local & 14 & Age-language match; stable personality; natural catchphrases and punctuation; occupation/education fit; role-specific social style. \\
D4 & Local & 8 & No excess expertise; natural uncertainty on weak topics; no unauthorized actions; responds from life experience rather than fake authority. \\
D5 & Local & 8 & Appropriate segmentation count; each segment is readable; segment order and reply rhythm are plausible. \\
E1 & Global & 14 & No repeated questions; facts stay coherent; speaker references are correct; state changes are tracked; memory is not confused with the present. \\
E2 & Global & 10 & Responds to the trigger event; respects relationship boundaries; tracks time and place; tone matches the scenario. \\
E3 & Global & 11 & Perceives emotion; comforts before solving when needed; intimacy is appropriate; no cold shutdown; boundaries are respected. \\
E4 & Global & 9 & Uses follow-up questions, self-disclosure, topic steering, or action proposals; avoids being fully passive. \\
E5 & Global & 4 & Has a natural trigger; both participants contribute; interaction is effective; the dialogue arc develops rather than stalls. \\
\bottomrule
\end{tabularx}
\end{table}

\subsection{Scoring, Acceptance, and Diagnostics}

Each dimension contains binary checkboxes. For a dimension $d$ with $m_d$ checkboxes, the score is
\[
  q_d = \frac{1}{m_d}\sum_{j=1}^{m_d} \mathbb{1}[b_{d,j}=\mathrm{true}].
\]
The benchmark then computes a weight-normalized local score and a weight-normalized global score:
\[
  S_{\mathrm{turn}} =
  \frac{1}{|T_r|}\sum_{t\in T_r}
  \frac{\sum_{d\in \mathcal{D}_{\mathrm{turn}}} w_d q_{t,d}}
       {\sum_{d\in \mathcal{D}_{\mathrm{turn}}} w_d},
  \qquad
  S_{\mathrm{dialogue}} =
  \frac{\sum_{d\in \mathcal{D}_{\mathrm{dialogue}}} w_d q_d}
       {\sum_{d\in \mathcal{D}_{\mathrm{dialogue}}} w_d}.
\]
Unless otherwise specified, the final score is
\[
  S = 100\cdot (0.5 S_{\mathrm{turn}} + 0.5 S_{\mathrm{dialogue}})
\]
for L0-valid cases. The benchmark also reports a strict \textbf{ACC}. A case receives ACC=1 only when it is L0-valid and every checkbox in every evaluated-role turn and in the holistic card passes. Model-level ACC values in the result tables are computed over L0-valid cases, while valid rate and L0 failure counts are tracked separately by the report generator over all scheduled cases. This separation matters because a model may be high quality when valid but still unreliable as a data generator.

The report generator aggregates valid rate, L0 failure count, ACC, final score, per-turn and holistic scores, per-dimension score, per-dimension ACC, category-level score, and the lowest-ACC failure dimensions. It also writes full \texttt{detail.json} records containing each failed checkbox and its judge rationale. The same trace format is used for benchmark comparison and data curation, so a failed case can be routed to rejection, repair, preference-pair construction, or regression testing.

\subsection{Benchmark Workflow}

The benchmark is executed as a fixed six-stage pipeline. First, persona-scenario bindings are expanded into role-conditioned cases. Second, the tested model and a reference partner generate a dialogue under the same scheduler. Third, the L0 gate determines whether the case is valid. Fourth, each valid evaluated-role turn is scored with the local rubric. Fifth, the whole dialogue is scored with the global rubric. Finally, the case-level trace is aggregated into model-level summaries. Table~\ref{tab:workflow} summarizes the protocol.

\begin{table}[t]
\centering
\small
\caption{Benchmark execution workflow. The evaluated role is the persona controlled by the tested model in the current case.}
\label{tab:workflow}
\begin{tabularx}{\linewidth}{@{}lX@{}}
\toprule
Stage & Operation \\
\midrule
Case construction & Load a binding $(p_a,p_b,s)$ and create two cases, one with the tested model as $p_a$ and one with it as $p_b$. \\
Dialogue generation & Instantiate both personas with the same scenario, opener, virtual start time, duration horizon, and single-draft scheduler; save dialogue and raw model-call traces. \\
L0 gate & Check deterministic identity, medium, co-presence, structure, and repetition failures; mark fatal cases invalid before soft scoring. \\
Local scoring & For each evaluated-role turn, call the judge with the target turn and a five-turn context window; compute D1--D5 checkbox scores. \\
Global scoring & Call the judge once with the full dialogue and evaluated role marker; compute E1--E5 checkbox scores. \\
Aggregation & Write eval traces, case details, model summaries, category breakdowns, strict ACC, and weakest dimensions by ACC. \\
\bottomrule
\end{tabularx}
\end{table}

\section{Experiments}

\subsection{Benchmark Setup}

The current benchmark uses the repository's fixed persona, scenario, and binding files. The corpus contains 55 persona cards, 50 scenario cards, and 50 persona-scenario bindings. Each binding specifies two personas and one scenario, and the runner expands it into two role-conditioned cases: one case evaluates the tested model as \texttt{persona\_a}, and the other evaluates it as \texttt{persona\_b}. Each model is therefore evaluated on 100 cases under identical persona-scenario coverage.

\begin{table}[t]
\centering
\small
\caption{Benchmark corpus and runtime configuration used for the current model comparison.}
\label{tab:benchmark_config}
\begin{tabularx}{\linewidth}{@{}lX@{}}
\toprule
Item & Value \\
\midrule
Personas & 55 role cards with identity, age, occupation, personality, expression habits, weak topics, and knowledge boundaries. \\
Scenarios & 50 scenario cards with relationship, trigger event, emotional tone, opener, virtual start time, and duration horizon. \\
Bindings & 50 fixed persona-scenario bindings, expanded into 100 evaluated-role cases per tested model. \\
Categories & Work, daily chat, emotional support, relationship development, interests, pets/family life, parenting, sports, finance, news, gaming, and technology. \\
Duration horizon & 8, 10, 15, 20, or 25 virtual minutes, depending on the scenario card. \\
Evaluated role & Only the persona controlled by the tested model is scored; the partner is generated by a fixed reference model. \\
Judge model & Qwen3.5-397B, run with zero-temperature scoring prompts for L0, local, and holistic evaluation. \\
\bottomrule
\end{tabularx}
\end{table}

In each case, the tested model controls one persona and the fixed reference partner controls the other. Both agents run on the same single-draft scheduler, with the same scenario opener, virtual time state, duration horizon, and maximum-turn configuration. This design evaluates whether the tested model can behave anthropomorphically in an interactive conversation rather than only produce a single isolated reply. The partner is not scored in that case; it supplies a stable interaction environment and prevents model comparisons from being confounded by changing both sides of the dialogue at once.

The benchmark records three levels of evidence. The generated dialogue is saved as a human-readable conversation artifact. Raw traces store each model call, including the assembled system prompt, user snapshot, previous draft state, and model output. Evaluation traces store L0 violations, every per-turn score card, the holistic score card, failed checkboxes, and judge rationales. These artifacts make it possible to audit individual failures, resume completed cases from cache, regenerate summaries without new judge calls, and construct accepted, rejected, or repaired dialogue examples.

The Qwen3.5-397B judge is run at zero temperature to reduce scoring variance. For valid cases, the runner scores all evaluated-role turns in parallel subject to an adaptive concurrency cap, while a separate holistic call scores the full dialogue. Per-turn scoring always uses the five-turn local window described in Table~\ref{tab:workflow}; holistic scoring sees the complete conversation. Each model summary reports valid rate, L0 failure count, strict ACC, final score, per-turn score, holistic score, per-dimension scores, per-dimension ACC, category scores, category ACC, per-case details, and the three weakest dimensions by ACC.

\subsection{Evaluated Systems and Inference Modes}

The current benchmark snapshot contains 16 systems and variants. We include all rows in the paper rather than only a representative subset. The comparison contains frontier chat baselines, open Qwen-family baselines, thinking-mode variants, no-think variants, and post-training variants produced by our pipeline. In the tables, \texttt{SFT} denotes supervised fine-tuning on selected anthropomorphic dialogue traces, while \texttt{SFT+RL} denotes continued CDT-ZPD guided GRPO. Rows marked \texttt{think} use explicit test-time thinking. Rows marked \texttt{no-think}, and rows without a \texttt{think} suffix, use no-think inference.

This distinction is especially important for the ablation rows. For readability, the result tables place general baselines first and group the 9B and 27B untuned/SFT/SFT+RL variants at the bottom. The 9B SFT and 9B SFT+RL rows are evaluated without thinking inference because this smaller model is intended as a fast-response system variant, where interactive latency and smooth chat rhythm are part of the design target. Their gains therefore measure whether the pipeline can transfer anthropomorphic behavior into a smaller no-think generation model. The final highlighted row, Qwen3.6-27B-SFT+RL, corresponds to the full 27B SFT+RL system with thinking inference.

\subsection{Current Model Comparison}

Table~\ref{tab:leaderboard} reports the complete benchmark snapshot. Among non-trained baselines, Gemini-3.5-Flash obtains the strongest strict ACC at 32.00\%, while GPT-5.5 obtains the strongest holistic ACC and continuous holistic score. The highlighted Qwen3.6-27B-SFT+RL system obtains 39.00\% strict ACC, 88.09\% turn ACC, and a 98.5 overall score. This improves strict all-checkbox acceptance by 7.00 absolute points over the strongest non-trained baseline and by 20.00 absolute points over the 27B SFT-only variant. In the efficiency-oriented 9B no-think setting, SFT raises strict ACC from 0.00\% to 13.00\%, and SFT+RL further raises it to 18.37\%. These results suggest that post-training contributes behavior that is not reducible to test-time thinking alone, while also making a lower-overhead deployment path viable.

\begin{table}[t]
\centering
\small
\caption{Complete benchmark results. Each model is scheduled on 100 role-conditioned cases; ACC values are strict all-checkbox acceptance rates over L0-valid cases. Best values are bolded and second-best values are underlined. The 9B and 27B ablations are grouped at the bottom; Qwen3.6-27B-SFT+RL is placed last and highlighted.}
\label{tab:leaderboard}
\resizebox{\linewidth}{!}{%
\begin{tabular}{lrrrrrr}
\toprule
Model & ACC & Turn ACC & Holistic ACC & Score & Per-turn & Holistic \\
\midrule
Gemini-3.5-Flash & \secondscore{32.00\%} & 81.87\% & \secondscore{87.00\%} & \secondscore{97.3} & 0.9614 & 0.9848 \\
GPT-5.5 & 25.00\% & 80.42\% & \bestscore{90.00\%} & 96.8 & 0.9460 & \bestscore{0.9905} \\
DeepSeek-V4-Pro & 22.00\% & 82.27\% & 71.00\% & 96.5 & 0.9645 & 0.9652 \\
Qwen3.5-397B-A17B (think) & 18.00\% & \secondscore{86.94\%} & 60.00\% & 96.3 & \secondscore{0.9811} & 0.9444 \\
Claude-Opus-4-8 & 10.00\% & 79.46\% & 52.00\% & 95.5 & 0.9700 & 0.9408 \\
Qwen3.5-397B-A17B (no-think) & 12.12\% & 84.75\% & 44.44\% & 94.8 & 0.9794 & 0.9173 \\
Qwen3.6-35B-A3B (think) & 3.03\% & 71.43\% & 36.36\% & 91.6 & 0.9340 & 0.8971 \\
Qwen3.6-35B-A3B (no-think) & 6.59\% & 74.82\% & 36.26\% & 90.0 & 0.9514 & 0.8494 \\
\midrule
Qwen3.5-9B (no-think) & 0.00\% & 54.84\% & 1.10\% & 74.3 & 0.8633 & 0.6223 \\
Qwen3.5-9B (think) & 1.00\% & 69.53\% & 18.00\% & 86.3 & 0.9370 & 0.7888 \\
Qwen3.5-9B-SFT (no-think) & 13.00\% & 78.68\% & 32.00\% & 91.9 & 0.9661 & 0.8713 \\
\rowcolor{highlightrow}
Qwen3.5-9B-SFT+RL (no-think) & 18.37\% & 83.61\% & 39.80\% & 92.4 & \bestscore{0.9826} & 0.8645 \\
\midrule
Qwen3.6-27B (no-think) & 7.29\% & 77.34\% & 32.29\% & 90.7 & 0.9608 & 0.8522 \\
Qwen3.6-27B (think) & 16.00\% & 74.87\% & 74.00\% & 95.5 & 0.9461 & 0.9637 \\
Qwen3.6-27B-SFT (think) & 19.00\% & 76.27\% & 84.00\% & 96.7 & 0.9468 & 0.9871 \\
\rowcolor{highlightrow}
Qwen3.6-27B-SFT+RL (think) & \bestscore{39.00\%} & \bestscore{88.09\%} & 83.00\% & \bestscore{98.5} & \bestscore{0.9826} & \secondscore{0.9875} \\
\bottomrule
\end{tabular}
}
\end{table}

\subsection{Per-Dimension Results}

Tables~\ref{tab:dim_scores} and~\ref{tab:dim_acc} give the per-dimension breakdown. The full tables show that local and global dimensions respond differently to post-training. The 9B SFT+RL no-think model reaches the best average scores on D1 naturalness, D2 concision, and D5 rhythm, but remains much weaker on global context and scenario dimensions. The full 27B SFT+RL system is strongest on D3 persona consistency, D4 boundary control, E1 context consistency, and tied best on E2 scenario fit. This supports the benchmark design: anthropomorphic quality cannot be inferred from local message fluency alone.

\begin{table}[t]
\centering
\scriptsize
\setlength{\tabcolsep}{3pt}
\caption{Per-dimension average scores for all evaluated systems. Scores are normalized to $[0,1]$.}
\label{tab:dim_scores}
\resizebox{\linewidth}{!}{%
\begin{tabular}{lrrrrrrrrrr}
\toprule
Model & D1 & D2 & D3 & D4 & D5 & E1 & E2 & E3 & E4 & E5 \\
\midrule
Gemini-3.5-Flash & 0.9661 & 0.9718 & 0.9624 & 0.9896 & 0.9694 & 0.9640 & \secondscore{0.9925} & \bestscore{0.9980} & 0.9880 & 0.9950 \\
GPT-5.5 & 0.9484 & 0.9594 & 0.9418 & 0.9761 & 0.9479 & \secondscore{0.9820} & \bestscore{0.9950} & \bestscore{0.9980} & 0.9900 & 0.9900 \\
DeepSeek-V4-Pro & 0.9710 & 0.9589 & 0.9612 & 0.9884 & 0.9416 & 0.8980 & 0.9900 & 0.9940 & \bestscore{0.9960} & 0.9900 \\
Qwen3.5-397B-A17B (think) & \secondscore{0.9852} & \secondscore{0.9885} & 0.9731 & 0.9872 & 0.9827 & 0.8400 & 0.9800 & \bestscore{0.9980} & 0.9880 & 0.9750 \\
Claude-Opus-4-8 & 0.9638 & 0.9857 & 0.9485 & 0.9851 & 0.9803 & 0.8560 & 0.9800 & 0.9940 & 0.9520 & 0.9675 \\
Qwen3.5-397B-A17B (no-think) & 0.9818 & 0.9858 & \secondscore{0.9741} & 0.9871 & \secondscore{0.9846} & 0.7657 & 0.9848 & 0.9838 & 0.9778 & 0.9596 \\
Qwen3.6-35B-A3B (think) & 0.9559 & 0.9005 & 0.9416 & 0.9745 & 0.8894 & 0.7071 & 0.9520 & 0.9879 & \secondscore{0.9939} & 0.9571 \\
Qwen3.6-35B-A3B (no-think) & 0.9502 & 0.9489 & 0.9334 & 0.9738 & 0.9485 & 0.6264 & 0.9313 & 0.9385 & 0.9736 & 0.9011 \\
\midrule
Qwen3.5-9B (no-think) & 0.8942 & 0.8370 & 0.8317 & 0.9553 & 0.8121 & 0.2000 & 0.7637 & 0.7890 & 0.8857 & 0.6951 \\
Qwen3.5-9B (think) & 0.9477 & 0.9409 & 0.9203 & 0.9722 & 0.9249 & 0.4820 & 0.8600 & 0.9220 & 0.9660 & 0.9200 \\
Qwen3.5-9B-SFT (no-think) & 0.9571 & 0.9876 & 0.9385 & 0.9854 & 0.9822 & 0.6960 & 0.9400 & 0.9700 & 0.9220 & 0.9275 \\
\rowcolor{highlightrow}
Qwen3.5-9B-SFT+RL (no-think) & \bestscore{0.9862} & \bestscore{0.9940} & 0.9662 & \secondscore{0.9901} & \bestscore{0.9931} & 0.7082 & 0.9260 & 0.9449 & 0.9184 & 0.9158 \\
\midrule
Qwen3.6-27B (no-think) & 0.9663 & 0.9700 & 0.9477 & 0.9844 & 0.9676 & 0.6667 & 0.9401 & 0.9333 & 0.9292 & 0.8854 \\
Qwen3.6-27B (think) & 0.9585 & 0.9373 & 0.9575 & 0.9741 & 0.9205 & 0.9020 & 0.9850 & 0.9920 & 0.9920 & 0.9850 \\
Qwen3.6-27B-SFT (think) & 0.9572 & 0.9366 & 0.9577 & 0.9795 & 0.9242 & 0.9680 & \secondscore{0.9925} & \bestscore{0.9980} & 0.9920 & \bestscore{1.0000} \\
\rowcolor{highlightrow}
Qwen3.6-27B-SFT+RL (think) & 0.9849 & 0.9879 & \bestscore{0.9813} & \bestscore{0.9910} & 0.9837 & \bestscore{0.9840} & \bestscore{0.9950} & \secondscore{0.9960} & 0.9700 & \secondscore{0.9975} \\
\bottomrule
\end{tabular}
}
\end{table}

\begin{table}[t]
\centering
\scriptsize
\setlength{\tabcolsep}{3pt}
\caption{Per-dimension strict ACC values for all evaluated systems. A dimension passes only when all of its checkboxes pass. Values are percentages.}
\label{tab:dim_acc}
\resizebox{\linewidth}{!}{%
\begin{tabular}{lrrrrrrrrrr}
\toprule
Model & D1 & D2 & D3 & D4 & D5 & E1 & E2 & E3 & E4 & E5 \\
\midrule
Gemini-3.5-Flash & 93.92\% & 94.24\% & 89.05\% & \secondscore{96.96\%} & 94.88\% & 91.00\% & \secondscore{97.00\%} & \bestscore{99.00\%} & 95.00\% & 98.00\% \\
GPT-5.5 & 90.68\% & 92.41\% & 87.40\% & 94.48\% & 92.32\% & \bestscore{96.00\%} & \bestscore{98.00\%} & \bestscore{99.00\%} & 95.00\% & 97.00\% \\
DeepSeek-V4-Pro & 94.01\% & 91.83\% & 88.38\% & 96.64\% & 91.92\% & 72.00\% & 96.00\% & \bestscore{99.00\%} & \bestscore{98.00\%} & 96.00\% \\
Qwen3.5-397B-A17B (think) & \bestscore{96.74\%} & \secondscore{97.54\%} & \secondscore{92.29\%} & 96.28\% & 97.07\% & 64.00\% & 92.00\% & \bestscore{99.00\%} & 94.00\% & 91.00\% \\
Claude-Opus-4-8 & 89.29\% & 97.02\% & 85.08\% & 96.14\% & 96.66\% & 66.00\% & 92.00\% & 97.00\% & 77.00\% & 90.00\% \\
Qwen3.5-397B-A17B (no-think) & 95.80\% & 96.88\% & 91.87\% & 95.26\% & 96.82\% & 49.49\% & 95.96\% & 95.96\% & 89.90\% & 84.85\% \\
Qwen3.6-35B-A3B (think) & 92.12\% & 82.28\% & 85.06\% & 92.88\% & 81.84\% & 38.38\% & 80.81\% & 95.96\% & 96.97\% & 85.86\% \\
Qwen3.6-35B-A3B (no-think) & 90.41\% & 90.14\% & 82.98\% & 92.78\% & 90.75\% & 38.46\% & 76.92\% & 85.71\% & 90.11\% & 71.43\% \\
\midrule
Qwen3.5-9B (no-think) & 81.12\% & 70.55\% & 65.53\% & 88.09\% & 69.75\% & 3.30\% & 35.16\% & 57.14\% & 58.24\% & 32.97\% \\
Qwen3.5-9B (think) & 89.42\% & 87.86\% & 80.84\% & 91.95\% & 87.21\% & 20.00\% & 53.00\% & 83.00\% & 87.00\% & 76.00\% \\
Qwen3.5-9B-SFT (no-think) & 89.96\% & 96.78\% & 83.27\% & 95.88\% & 96.27\% & 45.00\% & 81.00\% & 90.00\% & 73.00\% & 78.00\% \\
\rowcolor{highlightrow}
Qwen3.5-9B-SFT+RL (no-think) & 95.87\% & \bestscore{98.25\%} & 88.72\% & 96.58\% & \bestscore{98.57\%} & 48.98\% & 79.59\% & 85.71\% & 74.49\% & 77.55\% \\
\midrule
Qwen3.6-27B (no-think) & 92.88\% & 93.68\% & 86.28\% & 94.94\% & 93.61\% & 38.54\% & 82.29\% & 84.38\% & 79.17\% & 75.00\% \\
Qwen3.6-27B (think) & 90.75\% & 88.42\% & 88.76\% & 92.91\% & 87.64\% & 77.00\% & 94.00\% & 97.00\% & \bestscore{98.00\%} & 94.00\% \\
Qwen3.6-27B-SFT (think) & 91.97\% & 88.33\% & 89.10\% & 93.41\% & 87.57\% & 87.00\% & \secondscore{97.00\%} & \bestscore{99.00\%} & \secondscore{97.00\%} & \bestscore{100.00\%} \\
\rowcolor{highlightrow}
Qwen3.6-27B-SFT+RL (think) & \secondscore{96.38\%} & 97.21\% & \bestscore{93.97\%} & \bestscore{97.06\%} & \secondscore{97.36\%} & \secondscore{95.00\%} & \bestscore{98.00\%} & \secondscore{98.00\%} & 88.00\% & \secondscore{99.00\%} \\
\bottomrule
\end{tabular}
}
\end{table}

\subsection{Scenario Category Results}

The benchmark also reports category-level scores and category-level strict ACC across 12 scenario families. Table~\ref{tab:category_scores} shows continuous category scores, while Table~\ref{tab:category_acc} reports strict all-checkbox category acceptance. Qwen3.6-27B-SFT+RL obtains the best or second-best continuous score on most categories, with especially strong results on sports, relationship development, pet and family life, parenting, news, daily chat, and technology. The stricter category ACC table shows that some categories remain difficult even when continuous scores are high, especially technology and finance. This gap again supports reporting both graded quality and strict acceptance.

\begin{table}[t]
\centering
\scriptsize
\setlength{\tabcolsep}{2pt}
\caption{Category-level scores for all evaluated systems. Scores are on a 0--100 scale.}
\label{tab:category_scores}
\resizebox{\linewidth}{!}{%
\begin{tabular}{lrrrrrrrrrrrr}
\toprule
Model & Sport & Relation & Interest & Pet/Fam. & Parenting & Work & Emotion & News & Daily & Game & Tech & Finance \\
\midrule
Gemini-3.5-Flash & 99.3 & \bestscore{99.5} & \bestscore{98.3} & 95.6 & 97.4 & 94.5 & 96.8 & 97.4 & 97.6 & \bestscore{99.7} & 95.8 & 96.7 \\
GPT-5.5 & 98.8 & 97.8 & 97.2 & 97.0 & 97.3 & 92.0 & 94.8 & 95.6 & \secondscore{98.3} & 99.3 & 96.8 & \bestscore{98.7} \\
DeepSeek-V4-Pro & 98.0 & 96.2 & 97.2 & \secondscore{97.7} & 95.9 & 94.6 & \bestscore{98.7} & 94.3 & 97.3 & 99.1 & 94.7 & 93.9 \\
Qwen3.5-397B-A17B (think) & 97.9 & 94.1 & 95.5 & 93.4 & \secondscore{98.0} & 95.2 & \secondscore{98.2} & 96.8 & 96.5 & 96.0 & \secondscore{96.9} & 96.8 \\
Claude-Opus-4-8 & 97.0 & 96.7 & \secondscore{97.4} & 94.3 & 94.0 & \secondscore{96.9} & 96.2 & 94.5 & 97.0 & 96.6 & 91.0 & 92.6 \\
Qwen3.5-397B-A17B (no-think) & 95.3 & 91.0 & 95.3 & 95.3 & 93.8 & 93.8 & 98.1 & 92.5 & 95.7 & 98.4 & 92.8 & 95.3 \\
Qwen3.6-35B-A3B (think) & 95.7 & 88.6 & 94.7 & 88.0 & 85.9 & 92.6 & 93.7 & 92.5 & 93.0 & 91.6 & 94.3 & 87.0 \\
Qwen3.6-35B-A3B (no-think) & 94.8 & 93.8 & 91.5 & 91.0 & 91.2 & 89.3 & 91.4 & 87.8 & 89.6 & 88.0 & 78.8 & 90.3 \\
\midrule
Qwen3.5-9B (no-think) & 77.9 & 71.7 & 77.1 & 72.6 & 58.0 & 79.5 & 80.6 & 74.8 & 75.6 & 71.4 & 70.1 & 77.8 \\
Qwen3.5-9B (think) & 87.6 & 87.6 & 89.8 & 87.9 & 82.6 & 85.8 & 86.2 & 80.0 & 86.8 & 78.9 & 88.2 & 91.8 \\
Qwen3.5-9B-SFT (no-think) & 91.7 & 86.4 & 92.7 & 89.8 & 87.7 & 95.7 & 91.6 & 92.9 & 91.5 & 96.3 & 94.2 & 92.8 \\
\rowcolor{highlightrow}
Qwen3.5-9B-SFT+RL (no-think) & 93.1 & 90.9 & 96.7 & 86.1 & 83.0 & 96.7 & 91.9 & 93.8 & 95.1 & 92.6 & 88.5 & 96.9 \\
\midrule
Qwen3.6-27B (no-think) & 93.5 & 88.6 & 94.5 & 88.3 & 79.7 & 88.7 & 88.3 & 90.8 & 93.7 & 95.3 & 93.0 & 94.8 \\
Qwen3.6-27B (think) & \secondscore{99.4} & 95.6 & 95.9 & 93.8 & 94.3 & 92.9 & 97.6 & 93.7 & 96.3 & 98.2 & 92.8 & 95.1 \\
Qwen3.6-27B-SFT (think) & 99.3 & \secondscore{98.4} & 96.8 & 96.5 & \secondscore{98.0} & 94.0 & 97.2 & \secondscore{98.0} & 93.8 & 97.6 & 95.9 & 96.2 \\
\rowcolor{highlightrow}
Qwen3.6-27B-SFT+RL (think) & \bestscore{99.8} & \bestscore{99.5} & 97.2 & \bestscore{99.0} & \bestscore{99.8} & \bestscore{97.2} & 97.9 & \bestscore{98.6} & \bestscore{98.5} & \secondscore{99.6} & \bestscore{98.8} & \secondscore{97.4} \\
\bottomrule
\end{tabular}
}
\end{table}

\begin{table}[t]
\centering
\tiny
\setlength{\tabcolsep}{1.5pt}
\caption{Category-level strict ACC for all evaluated systems. ACC is computed over valid cases only.}
\label{tab:category_acc}
\resizebox{\linewidth}{!}{%
\begin{tabular}{lrrrrrrrrrrrrr}
\toprule
Model & ACC & Sport & Relation & Interest & Pet/Fam. & Parenting & Work & Emotion & News & Daily & Game & Tech & Finance \\
\midrule
Gemini-3.5-Flash & \secondscore{32.00\%} & 37.5\% & \secondscore{25.0\%} & \secondscore{30.0\%} & \bestscore{37.5\%} & 12.5\% & \bestscore{30.0\%} & \secondscore{30.0\%} & 25.0\% & 30.0\% & \bestscore{66.7\%} & \bestscore{50.0\%} & \bestscore{25.0\%} \\
GPT-5.5 & 25.00\% & \bestscore{50.0\%} & 12.5\% & 20.0\% & 12.5\% & \secondscore{25.0\%} & \bestscore{30.0\%} & 0.0\% & 25.0\% & 40.0\% & \secondscore{50.0\%} & \secondscore{33.3\%} & \secondscore{12.5\%} \\
DeepSeek-V4-Pro & 22.00\% & 25.0\% & 0.0\% & \secondscore{30.0\%} & \bestscore{37.5\%} & 0.0\% & \secondscore{20.0\%} & \secondscore{30.0\%} & 25.0\% & 30.0\% & 33.3\% & 0.0\% & \bestscore{25.0\%} \\
Qwen3.5-397B-A17B (think) & 18.00\% & 12.5\% & 12.5\% & \bestscore{40.0\%} & 12.5\% & 0.0\% & \bestscore{30.0\%} & 20.0\% & 12.5\% & 20.0\% & 0.0\% & 16.7\% & \bestscore{25.0\%} \\
Claude-Opus-4-8 & 10.00\% & 0.0\% & 0.0\% & 10.0\% & 12.5\% & 12.5\% & 10.0\% & 10.0\% & 12.5\% & 20.0\% & 33.3\% & 0.0\% & 0.0\% \\
Qwen3.5-397B-A17B (no-think) & 12.12\% & \secondscore{42.9\%} & 0.0\% & 20.0\% & 0.0\% & 0.0\% & \secondscore{20.0\%} & 10.0\% & 0.0\% & 20.0\% & 33.3\% & 0.0\% & 0.0\% \\
Qwen3.6-35B-A3B (think) & 3.03\% & 0.0\% & 0.0\% & 10.0\% & 0.0\% & 0.0\% & 10.0\% & 0.0\% & 0.0\% & 10.0\% & 0.0\% & 0.0\% & 0.0\% \\
Qwen3.6-35B-A3B (no-think) & 6.59\% & 12.5\% & 16.7\% & 10.0\% & 0.0\% & 0.0\% & 0.0\% & 0.0\% & 0.0\% & 10.0\% & 16.7\% & 0.0\% & \secondscore{12.5\%} \\
\midrule
Qwen3.5-9B (no-think) & 0.00\% & 0.0\% & 0.0\% & 0.0\% & 0.0\% & 0.0\% & 0.0\% & 0.0\% & 0.0\% & 0.0\% & 0.0\% & 0.0\% & 0.0\% \\
Qwen3.5-9B (think) & 1.00\% & 0.0\% & 0.0\% & 10.0\% & 0.0\% & 0.0\% & 0.0\% & 0.0\% & 0.0\% & 0.0\% & 0.0\% & 0.0\% & 0.0\% \\
Qwen3.5-9B-SFT (no-think) & 13.00\% & 0.0\% & 12.5\% & \secondscore{30.0\%} & 0.0\% & 0.0\% & \bestscore{30.0\%} & 10.0\% & 0.0\% & 20.0\% & 33.3\% & 0.0\% & \secondscore{12.5\%} \\
\rowcolor{highlightrow}
Qwen3.5-9B-SFT+RL (no-think) & 18.37\% & 25.0\% & 12.5\% & \secondscore{30.0\%} & 12.5\% & 0.0\% & 0.0\% & 10.0\% & \secondscore{28.6\%} & \secondscore{60.0\%} & 33.3\% & 0.0\% & 0.0\% \\
\midrule
Qwen3.6-27B (no-think) & 7.29\% & 0.0\% & 14.3\% & 22.2\% & 14.3\% & 0.0\% & 10.0\% & 0.0\% & 0.0\% & 11.1\% & 0.0\% & 16.7\% & 0.0\% \\
Qwen3.6-27B (think) & 16.00\% & 12.5\% & 0.0\% & \bestscore{40.0\%} & \secondscore{25.0\%} & 12.5\% & 10.0\% & 0.0\% & 25.0\% & 40.0\% & 0.0\% & 0.0\% & \secondscore{12.5\%} \\
Qwen3.6-27B-SFT (think) & 19.00\% & 25.0\% & 12.5\% & \secondscore{30.0\%} & 12.5\% & \secondscore{25.0\%} & 10.0\% & 20.0\% & 25.0\% & 30.0\% & 16.7\% & 0.0\% & \secondscore{12.5\%} \\
\rowcolor{highlightrow}
Qwen3.6-27B-SFT+RL (think) & \bestscore{39.00\%} & \bestscore{50.0\%} & \bestscore{50.0\%} & 20.0\% & \secondscore{25.0\%} & \bestscore{62.5\%} & \bestscore{30.0\%} & \bestscore{40.0\%} & \bestscore{50.0\%} & \bestscore{70.0\%} & \secondscore{50.0\%} & 0.0\% & \secondscore{12.5\%} \\
\bottomrule
\end{tabular}
}
\end{table}

\FloatBarrier
\section{Analysis and Discussion}

\paragraph{Local fluency and global anthropomorphic quality diverge.}
The benchmark shows that high per-turn naturalness does not guarantee a convincing conversation. The 9B SFT+RL no-think model obtains the strongest average scores on several local dimensions, including D1, D2, and D5, but still lags behind on global context, scenario fit, social-emotional fit, and dialogue arc. Models can generate short, natural, persona-compatible messages while losing track of earlier facts or failing to move the conversation forward. This supports the decision to evaluate both local and global dimensions.

\paragraph{Post-training changes behavior beyond test-time thinking.}
The ablations suggest that SFT and RL contribute learned anthropomorphic behavior rather than merely exposing it through longer inference. On 27B thinking inference, SFT alone reaches 19.00\% strict ACC, while SFT+RL reaches 39.00\%. In the 9B no-think inference setting, the untuned model obtains 0.00\% strict ACC, SFT reaches 13.00\%, and SFT+RL reaches 18.37\%. This 9B setting is deployment-motivated: it removes explicit test-time reasoning to support faster interactive response, then asks whether training can recover anthropomorphic behavior inside the policy. The 9B SFT+RL no-think model also exceeds the untrained Qwen3.6-27B thinking variant in strict ACC, indicating that post-training can compensate for some scale and inference-mode disadvantages.

\paragraph{Proactivity is not simply asking more questions.}
The E4 results show that anthropomorphic proactivity is subtle. A model can continue a conversation by sharing a small personal reaction, lightly teasing, shifting to a related topic, proposing an action, or closing gracefully. Repeatedly asking what happened next is often a sign of weak dialogue planning rather than strong proactivity.

\paragraph{Validity and quality should be reported separately.}
Several models have high scores among valid samples but low valid rates. This means the model can sound human-like when it avoids hard failures, yet remains unreliable as a data generator. For synthetic data construction, L0 valid rate is therefore as important as the average score.

\paragraph{Category ACC exposes remaining hard cases.}
Continuous category scores are high for many models, but strict category ACC remains low in several families. For example, the full system achieves strong continuous scores on technology and finance, but strict category ACC remains 0.0\% and 12.5\%, respectively. These failures are useful because they identify scenario families where a conversation can look good on average while still missing at least one required anthropomorphic constraint.

\paragraph{The first instantiated medium is culturally specific.}
The current benchmark is built around Chinese private-message behavior. This is useful for controlled evaluation because medium-specific norms make failures observable. The same design separates medium-specific checks from general behavioral dimensions, so the pipeline can preserve the gate-and-score structure while changing the surface norms used by the judge.

\section{Limitations}

The current results evaluate the implemented pipeline and benchmark snapshot rather than a deployed product. Automatic judges may be biased by length, position, style, and model family, so the reported scores should be interpreted as structured behavioral measurements rather than definitive human preference labels. Synthetic self-chat may reduce diversity if filtering is too strict or if skill evolution overfits to benchmark artifacts. The training ablations also mix model scale and inference mode: the 27B full system uses thinking inference, while the 9B SFT/RL ablations use no-think inference as an efficiency-oriented setting for faster interactive response. This version does not report latency measurements, so the 9B no-think results should be read as a quality evaluation of the low-overhead track rather than a full serving benchmark. Finally, the current benchmark is medium-specific, and the strongest claims in this version concern Chinese private-message dialogue.

\section{Ethics and Broader Impacts}

Anthropomorphic dialogue systems raise risks of deception and over-trust. Our formulation treats anthropomorphism as text behavior, not as evidence of consciousness or genuine emotion. A deployed system should disclose that the interlocutor is an AI system and should avoid implying human identity. This is especially important in emotionally vulnerable, medical, legal, financial, and other high-stakes settings. The evaluation rubric includes boundary failures to reduce overconfident expert behavior, but deployment policy must remain separate from benchmark optimization. Prior work has warned that fluent language systems can encourage misplaced attribution of understanding \citep{Weizenbaum1966ELIZA,Bender2021Dangers}. The purpose of this work is to measure and control human-like conversational behavior, not to hide system identity from users.

\section{Conclusion}

We introduced \methodname, a closed-loop framework for anthropomorphic dialogue generation, evaluation, and alignment. The framework operationalizes human-like private dialogue as observable behaviors involving persona, relationship, memory, boundary, rhythm, and conversation arc. The current pipeline connects persona-scenario conditioning, single-draft scheduling, memory compression, virtual time, rubric-based filtering, failure attribution, skill evolution, SFT, CDT-ZPD guided GRPO, and benchmark reporting. The benchmark snapshot shows that the full 27B SFT+RL system improves strict anthropomorphic acceptance over the strongest non-trained baseline, while the 9B no-think ablations show that SFT and RL also transfer useful anthropomorphic behavior into a smaller, faster-response model track without test-time thinking.

\bibliographystyle{plainnat}
\bibliography{references}

\end{document}